\documentclass[runningheads]{llncs}
\usepackage{xspace}
\usepackage{algorithm}
\usepackage[noend]{algorithmic}
\usepackage{amsmath,amsfonts}
\usepackage{mathrsfs}
\usepackage{enumitem}
\usepackage{bm}
\usepackage{multirow}
\usepackage{booktabs}
\usepackage{makecell}
\usepackage{graphicx}
\usepackage{caption}
\usepackage{lmodern} 
\usepackage{bbm}
\usepackage{bm}
\usepackage{ulem}
\usepackage{comment}
\usepackage[usenames,dvipsnames,svgnames]{xcolor} 
\usepackage{float}
\usepackage{subcaption}
\newcommand{\defeq}{\mathrel{\mathop:}=}

\newcommand{\vect}[1]{\ensuremath{\mathbf{#1}}}

\newcommand{\grad}{\nabla}
\newcommand{\hess}{\nabla^2}
\newcommand{\argmin}{\mathop{\rm argmin}}

\newcommand*{\minimize}{\mathrm{minimize}}

\newcommand{\norm}[1]{\|{#1}\|}

\newcommand{\trans}{^{\top}}

\newcommand{\proj}{\mathcal{P}}

\newcommand{\R}{\mathbb{R}}

\newcommand{\e}{\vect{e}}
\renewcommand{\u}{\vect{u}}
\renewcommand{\v}{\vect{v}}
\newcommand{\w}{\vect{w}}
\newcommand{\x}{\vect{x}}
\newcommand{\y}{\vect{y}}
\newcommand{\z}{\vect{z}}

\renewcommand{\H}{\mathcal{H}}

\newcommand{\cn}{\kappa}

\newcommand{\ca}{\hat{c}}
\newcommand{\Ts}{T'}
\newcommand{\Tt}{T''}

\newcommand{\ugrad}{\mathscr{G}}
\newcommand{\ufun}{\mathscr{F}}
\newcommand{\uspace}{\mathscr{S}}
\newcommand{\utime}{\mathscr{T}}

\renewcommand{\S}{\mathcal{S}}

\newcommand{\logterms}{\frac{d\cn}{\delta}}

\newcommand{\cXs}{\mathcal{X}_{\text{stuck}}}

\newcommand{\ball}{\mathbb{B}}

\newcommand{\ESSP}{$\varepsilon$-second-order stationary point\xspace}

\newcommand{\diff}{\mathrm{d}}
\newcommand{\dis}{\displaystyle}
\newtheorem{assumption}{Assumption}

\newcommand{\<}{\langle}
\renewcommand{\>}{\rangle}
\newcommand{\n}{\boldsymbol{n}}
\newcommand{\prox}{\mathrm{prox}}

\newcommand{\Lip}{L}  

\newenvironment{shortversion}{}{}
\newenvironment{longversion}{}{}

\excludecomment{shortversion}

\begin{document}

\title{Perturbed Proximal Descent to Escape Saddle Points for Non-convex and Non-smooth Objective Functions}
\titlerunning{Perturbed Proximal Descent}
\author{Zhishen Huang\inst{1} \and Stephen Becker\inst{1}\orcidID{0000-0002-1932-8159}}
\authorrunning{Z. Huang and S. Becker}
\institute{Dept.\ of Applied Math., University of Colorado, Boulder, USA
\email{\{zhishen.huang,stephen.becker\}@colorado.edu}}

\maketitle              
\begin{abstract}
We consider the problem of finding local minimizers in non-convex and non-smooth optimization. Under the assumption of strict saddle points, positive results have been derived for first-order methods. We present the first known results for the non-smooth case, which requires different analysis and a different algorithm.
\begin{longversion}
{\it  This is the extended version of the paper that contains the proofs.}
\end{longversion}
\end{abstract}

\section{Introduction}
We consider the problem of finding approximate local minimizers of the
problem
%
\begin{equation} \label{eq:main}
\minimize_{\x\in\R^d } \left( \Phi(\x) \defeq f(\x) + g(\x) \right)
\end{equation}
where $f(\x)$ is not convex but smooth (and with full domain), and $g(\x)$ is convex but not smooth. Many optimization problems in engineering, signal processing and machine learning can be cast in this framework, where $f$ is a smooth loss function, and $g$ is a non-smooth regularizer such as a norm. 
For example, our model captures regularized neural networks~\cite{GirosiPoggio95}, where the regularization can induce sparsity as an alternative to dropout. 
In this paper, for simplicity we restrict our discussion to $g(\x)=\lambda\|\x\|_1$, where $\lambda\ge 0$ is a constant, but many of the results apply to more general choices of $g$.
%
The first-order condition is $0 \in \nabla f(\x) + \partial g(\x)$, and any $\x$ satisfying this condition is called a ``stationary point" (see \cite{CombettesBook2} for background on the subdifferential $\partial g$). All local minimizers are stationary points, but not vice-versa.   We define a ``saddle point'' to be any stationary point where the Hessian is indefinite (and therefore not a local minimizer).  This paper extends a recent line of work \cite{jin2017escape} to analyze when we can expect to find a local minimizer.  It has been argued that in many machine learning problems, finding any local minimizer is often enough for good performance, but finding a saddle point is not useful~\cite{dauphin2014identifying}.

The fact that $g$ is non-smooth is crucially important, and it does more than just complicate the analysis, as it also requires a new algorithm. 
In the smooth case, $f$ is often minimized using gradient descent or an accelerated variant~\cite{Nesterov83} with a fixed stepsize. Na\"ively extending gradient descent to apply to \eqref{eq:main} leads to subgradient descent with fixed-stepsize. Unfortunately, this method fails to converge as the example $d=1, \lambda=1$ and $f=0$ shows~\cite{shor1962application} since for a generic choice of the initial point, the sequence is not Cauchy.

Instead of gradient descent, we use a perturbed version of proximal gradient descent. For a real-valued convex lower semi-continuous function $g$, define the ``proximity'' operator (or ``prox'' for short) as the map $\prox_{g}(\y) = \argmin_{\x} g(\y) + \frac{1}{2}\|\x-\y\|^2$ (throughout the paper, for vectors we use $\|\cdot\|$ to denote the Euclidean norm). Equivalently, $\prox_g = (I + \partial g)^{-1}$, and thus the first-order condition is equivalent to $\x = \prox_{\eta g}[ \x - \eta \nabla f(\x) ]$ for any $\eta>0$. Proximal gradient descent is the iteration $\x_{t+1} = \prox_{\eta g}[\x_{t} - \eta \nabla f(\x_t)]$, so it immediately follows that if the sequence converges, it converges to a stationary point. Convergence of the sequence is known to follow from mild assumptions on $f$ and $g$, the stepsize $\eta$, and boundedness of the sequence $\{\x_t\}$~\cite{attouch2011convergence}.

We define a {\it second-order stationary point} to be a first-order stationary point $\x$ that additionally satisfies $\nabla^2 f(\x) \succ 0$, which is a sufficient condition for $\x$ to be a local minimizer.
Our main contribution is showing that under suitable assumptions, a perturbed version of proximal gradient descent will generate a sequence that converges to an approximate second-order stationary point.
We make assumptions on the second-order behavior of $f$, similar to assumptions under which it is known that gradient descent will always converge to a second-order stationary point except for adversarially chosen starting points~\cite{lee2016gradient} --- in contrast to Newton's method, which is attracted to all stationary points. However, even in the smooth case when the sequence converges, gradient descent converges arbitrarily slowly~\cite{Du2017Octopus} in the presence of a saddle point, so perturbation is necessary. In the non-smooth case, perturbation is even more important due to the proximal nature of the algorithm.









\begin{figure} 
\begin{minipage}{.44\textwidth}
\centering
\includegraphics[width=\linewidth]{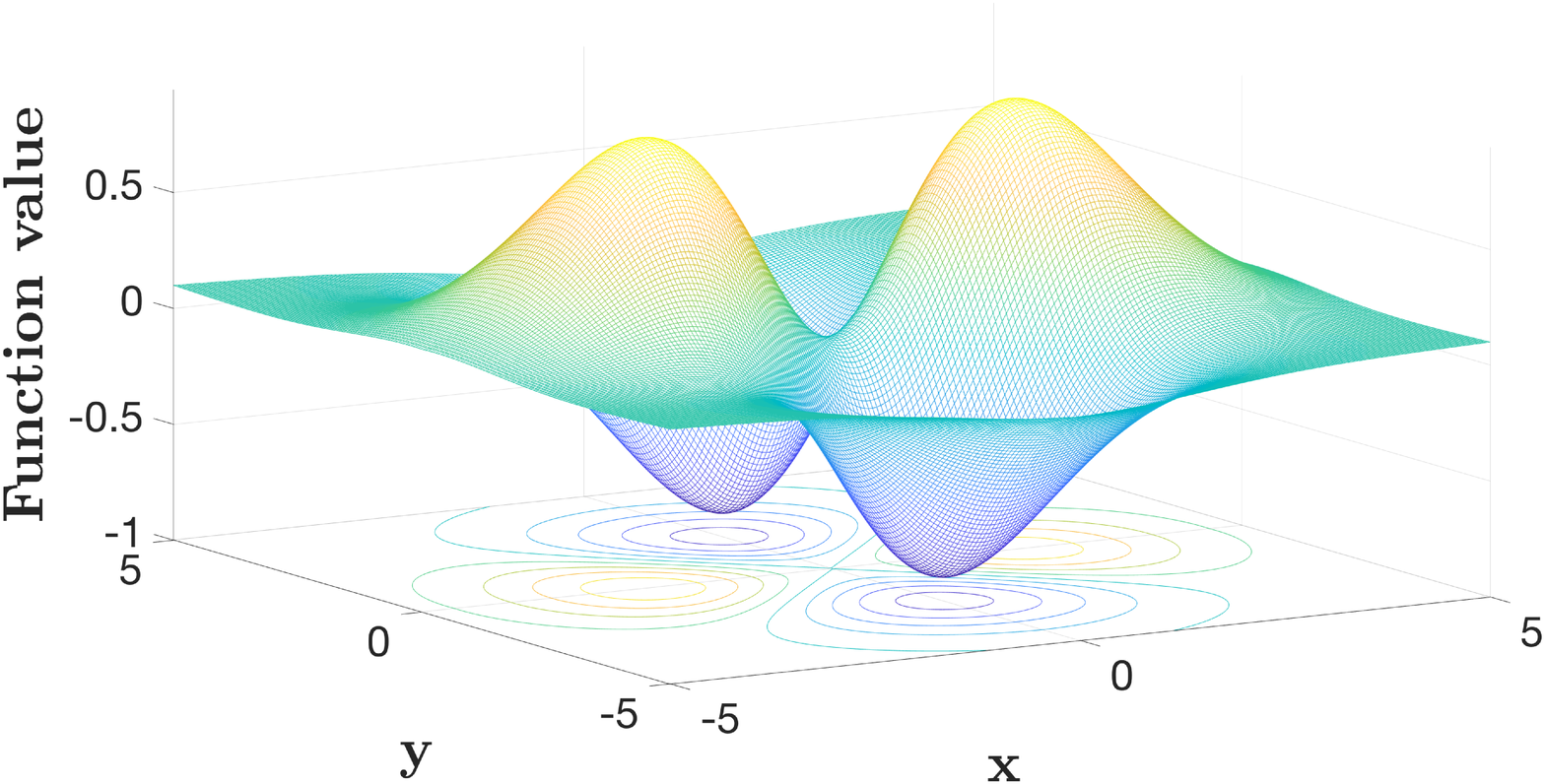}
\caption{Graph of function $\Phi(\x)$}
\label{fig::gaussian bump}
\end{minipage}
\begin{minipage}{.52\textwidth}
\centering
\includegraphics[width=.95\linewidth]{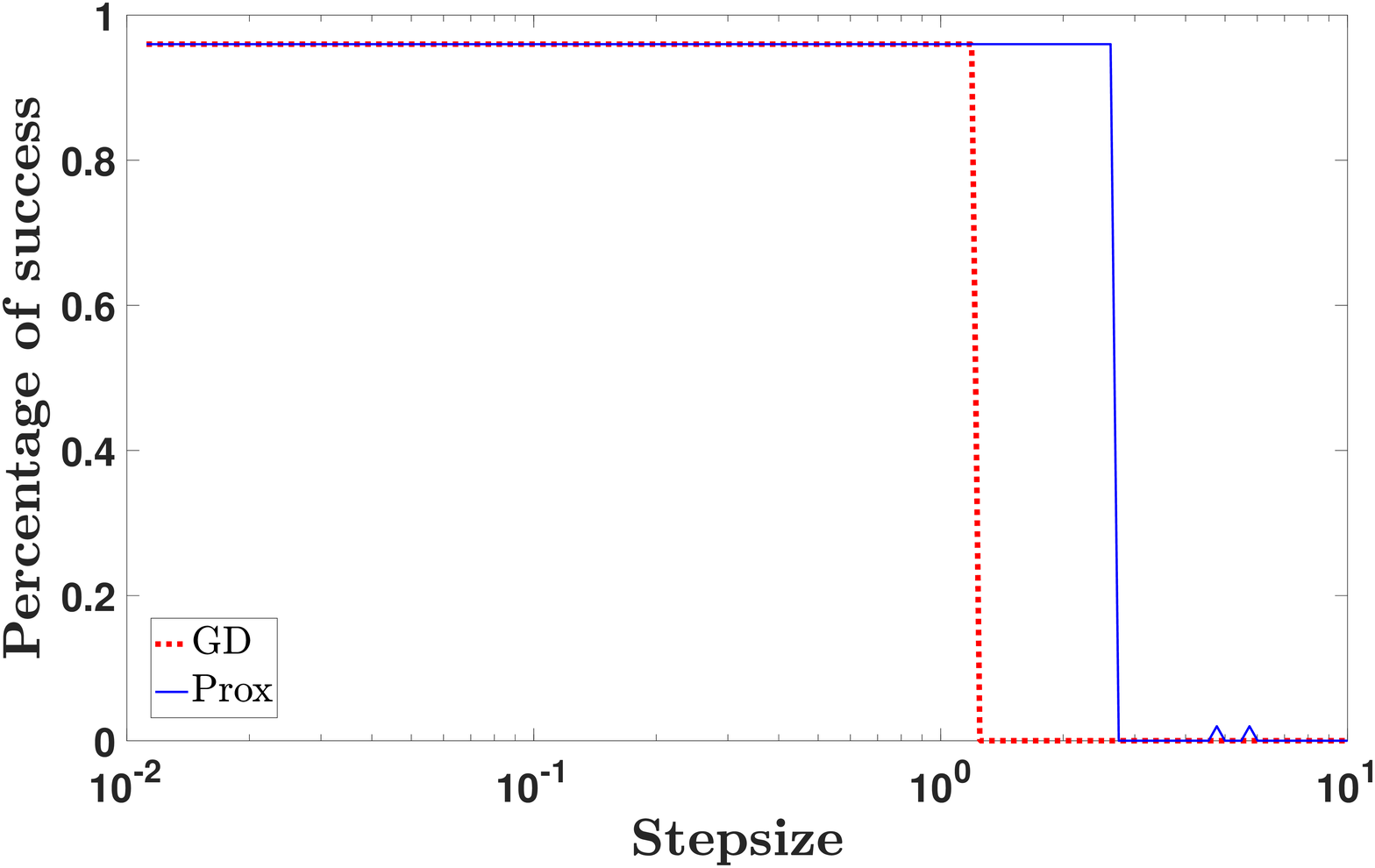}
\caption{The comparison between gradient descent (GD) and proximal gradient descent (Prox) on the percentage of success finding the correct local minima, as a function of the stepsize $\eta$}
\label{fig::succVSstepsize}
\end{minipage}
\end{figure}

\paragraph{A toy example: Gaussian Bump}
Consider the function $\Phi: \R^2 \rightarrow \R, 
x \mapsto \frac{1}{2}(x^2-y^2)\mathrm{e}^{-\frac{x^2+y^2}{5}}+\frac{1}{100}h_{100}(\x)
$
where $h_{100}(\x)$ is the Huber function with parameter 100 \cite{BeckBook2017}. The choice of this combination of Huber parameter and the magnitude of Huber function ensures that the origin is a saddle point. The Huber function approximates the $\ell_1$ norm. The plot is show in Fig.~\ref{fig::gaussian bump}.

This function
has two local minima and a saddle point at $(0,0)$.  Because the Huber function is both smooth and it has a known proximity operator, we can treat it as either part of the smooth $f$ component or the non-smooth $g$ component, and therefore run either gradient descent or proximal gradient descent. We experiment with both algorithms, randomly picking initial points at $\x_0=(0.3,0.01)+\boldsymbol{\xi}$ where $\boldsymbol{\xi}$ is sampled uniformly from $\mathbb{B}_0(\frac{1}{10}\|\x_0\|)$, and varying the stepsize $\eta$, with fixed maximum iteration 1000.
Figure \ref{fig::succVSstepsize} shows the empirical success rate of finding a local minimizer (as opposed to converging to the saddle point at $(0,0)$).


We observe that the range of stable step size for the proximal descent algorithm is wider than gradient descent, and the success rate of proximal descent is as high as  the gradient descent. This example motivates us to adopt proximal descent over gradient descent in real application for better stability and equivalent, if not better, accuracy.

\paragraph{A coincidence} In this toy example, the saddle point at $(0,0)$ happens to be a fixed point of proximal operator of $\eta\lambda\|\x\|_1$. Soft thresholding, as the proximal operator of $\lambda \|\x\|_1$ is known~\cite{CombettesWajs05}, has an attracting region that sets nearby points to $0$. The radius of the attracting region (per dimension) 
is $\eta\lambda$, thus if  $\|\x_{t_0}-\eta\nabla f(\x_{t_0})\|_\infty \le \eta\lambda$ for some iteration $t_0$, then $\x_t = 0$ for all $t>t_0$. Proximal gradient descent performs even better when the saddle point is not in the attracting region.


\begin{longversion}
\paragraph{Structure of the paper} Section \ref{sec::algorithm} states the algorithm, followed by section \ref{sec::main} where the theoretical guarantee is presented with proof. Section \ref{sec::numerics} shows numerical experiments.
\end{longversion}

\subsection{Related literature}

\paragraph{Second order methods for smooth objectives} 
Some recent second order methods, mainly based on either cubic-regularized Newton methods as in \cite{Nesterov2006CubicRugularise} or based on trust-region methods (as in Curtis et al.\ \cite{Curtis2017TrustRegion}), have been shown to converge to $\varepsilon$-approximate local minimizers of smooth non-convex objective functions in $\mathcal{O}(\varepsilon^{-1.5})$ iterations. See \cite{Carmon2017NoncvxOpt,jin2017escape,Zhu_Neon2} for a more thorough review of these methods. We do not consider these methods further due to the high-cost of solving for the Newton step in large dimensions.


\paragraph{First order methods for smooth objectives} We focus on first order methods because each step is cheaper and these methods are more frequently adopted by the deep learning community. Xu et al.\ in \cite{XuJinYang_NC_Extract} and Allen-Zhu et al.\ in \cite{Zhu_Neon2} develop Negative-Curvature (NC) search algorithms, which find descent direction corresponding to negative eigenvalues of Hessian matrix. The NC search routines avoid using either Hessian or Hessian-vector information directly, and it can be applied in both online and deterministic scenarios. In the online setting, combining NC search routine with first-order stochastic methods will give algorithms NEON-$\mathcal{A}$~\cite{XuJinYang_NC_Extract} and NEON2+SGD~\cite{Zhu_Neon2} with iteration cost  $\mathcal{O}(\frac{d}{\varepsilon^{3.5}})$ and $\mathcal{O}(\varepsilon^{-3.5})$ respectively (the latter still depends on dimension, whose induced complexity is at least $\ln^2(d)$), and these methods generate a sequence that converges to an approximate local minimum with high probability. In the offline setting, Jin et al.\ in \cite{jin2017escape} provide a stochastic first order method that finds an approximate local minimizer with high probability at computational cost $\mathcal{O}(\frac{\ln^4(d)}{\varepsilon^2})$. Combining NEON2 with gradient descent or SVRG, the cost to find an approximate local minimum is $\mathcal{O}(\varepsilon^{-2})$, whose dependence on dimension is not specified but at least $\ln^2(d)$.  These methods make Lipschitz continuity assumptions about the gradient and Hessian, so they do not apply to non-smooth optimization. 

A recent preprint \cite{liu2018envelope} approaches the problem of finding local minima using the forward-backward envelope technique developed in \cite{stella2017forward}, where the assumption about the smoothness of objective function is weakened to local smoothness instead of global smoothness. 


\paragraph{Non-smooth objectives} In the offline settings, Bo\c{t} et al.\ propose a proximal algorithm for minimizing non-convex and non-smooth objective functions in \cite{Bot2018proximal}. They show the convergence to KKT points instead of approximate second-order stationary points. Other work \cite{attouch2011convergence,Bolte13} relies on the Kurdya-Lojasiewicz inequality and shows convergence to stationary points in the sense of the limiting subdifferential, which is not the same as a local minimizer or approximate second-order stationary point. 
In the online setting, Reddi et al.\ demonstrated in \cite{Reddi2016proxSVRG} that the proximal descent with variance reduction technique (proxSVRG) has linear convergence to a first-order stationary point, but not to a local minimizer.

\section{Algorithm}
\label{sec::algorithm}

The algorithm takes as input a starting vector $\x_0$, the gradient Lipschitz constant $\Lip$, the    Hessian Lipschitz constant $\rho$,
   the second-order stationary point tolerance $\varepsilon$,
   a positive constant $c$, a
   failure probability $\delta$, and
   estimated function value gap $\Delta_\Phi$.
The key parameter for Algorithm \ref{algo::PPD} is the constant $c$. It should be made large enough so that the effect of perturbation will be significant enough for escaping saddle points, and at the same time not too large so that the iteration stepsize is of reasonable magnitude and the iteration will not go wild.
The output of the algorithm is an \ESSP (see Def.~\ref{def:essp}).


\begin{algorithm} 
\caption{Perturbed Proximal Descent: $\text{input}(\x_0, \Lip, \rho, \varepsilon, c, \delta, \Delta_\Phi)$}\label{algo::PPD}
\begin{algorithmic}
\STATE $\chi \leftarrow 3\max\{\ln(\frac{d\Lip\Delta_\Phi}{c\varepsilon^2\delta}), 4\},
 ~\eta \leftarrow \frac{c}{\Lip},
 ~r \leftarrow \frac{\sqrt{c}}{\chi^2}\cdot\frac{\varepsilon}{\Lip},
 ~g_{\text{thres}} \leftarrow \frac{\sqrt{c}}{\chi^2}\cdot \varepsilon,
 ~\Phi_{\text{thres}} \leftarrow \frac{c}{\chi^3} \cdot \sqrt{\frac{\varepsilon^3}{\rho}},
 ~t_{\text{thres}} \leftarrow \frac{\chi}{c^2}\cdot\frac{\Lip}{\sqrt{\rho \varepsilon}}$
\STATE $t_{\text{noise}} \leftarrow -t_{\text{thres}}-1$
\FOR{$t = 0, 1, \ldots $}
\IF{$\|\x - \mathrm{prox}_{\eta g}[\x-\eta\nabla f(\x)] \| < g_{\textrm{thres}} $ and $t - t_{\text{noise}} > t_{\text{thres}}$}
\STATE $\tilde{\x}_t \leftarrow \x_t, \quad t_{\text{noise}} \leftarrow t$
\STATE $\x_t \leftarrow \tilde{\x}_t + \boldsymbol{\xi}_t, \qquad \boldsymbol{\xi}_t \text{~uniformly~} \sim \ball_0(r)$
\ENDIF
\IF{$t - t_{\text{noise}} = t_{\text{thres}}$ and $\Phi(\x_t) - \Phi(\tilde{\x}_{t_{\text{noise}}}) > - \Phi_{\text{thres}}$}
\STATE \textbf{return} $\tilde{\x}_{t_{\text{noise}}}$
\ENDIF
\STATE $\x_{t+1} \leftarrow \prox_{\eta g}[\x_t - \eta \grad f(\x_t)]$
\ENDFOR
\end{algorithmic}
\end{algorithm}


\section{Escaping Saddle Points through Perturbed Proximal Descent}
\label{sec::main}
The main step in the algorithm is a proximal gradient descent step applied to $f+g$, defined as
\begin{align}  \label{eqn::proximal_descent}
\x_{t+1} &= \argmin_{\y} f(\x_t) + \<\nabla f(\x_t),\y-\x_t\> + \frac{\eta^{-1}}{2}\|\y-\x_t\|^2 + g(\y) \notag \\
&= \prox_{\eta g} \circ ( I-\eta\nabla f )(\x_t)
\end{align}


One motivation of preferring proximal descent to gradient descent, as shown in Figure \ref{fig::succVSstepsize}, is the stability of the algorithm with respect to stepsize change. The proximal step is similar to the implicit/backward Euler scheme, as equation \eqref{eqn::proximal_descent} can be written as 
$\x_{t+1} = \x_t - \eta\big(\nabla f(\x_t) + \partial g(\x_{t+1})\big)$. From this perspective, we expect that proximal descent will demonstrate at least the same convergence speed as gradient descent and stronger stability with respect to hyperparameter setting. 
\begin{definition}[Gradient Mapping]
Consider a function $\Phi(\x) = f(\x) + g(\x)$. The gradient mapping is defined as $G^{f,g}_{\eta}(\x) \defeq \x - \prox_{\eta g}[\x-\eta\nabla f(\x)]$
\end{definition}
In the rest of this paper, the super- and subscript of the gradient mapping are not specified, as it is always clear that $f$ represents the smooth nonconvex part of $\Phi$, $g$ represents $\lambda\|\x\|_1$, and $\eta$ is the stepsize used in the algorithm. Observe that the gradient map is just the gradient of $f$ if $g\equiv 0$.


\begin{definition}[First order stationary points]
For a function $\Phi(\x)$, define first order stationary points as the points which satisfy $ G(\x) = 0. $
\end{definition}

\begin{definition}[\ESSP] \label{def:essp}
Consider a function $\Phi(\x) = f(\x) + g(\x)$. 
A point $\x$ is an \ESSP if 
\begin{equation}
    \|G(\x)\| \le \varepsilon ~\textrm{ and }~ \lambda\big(\nabla^2 f(\x)\big)_{\min} \ge -\sqrt{\rho\varepsilon}
\end{equation}
where $\lambda(\cdot)_{\min}$ is the smallest eigenvalue.
\end{definition}

The first Lipschitz assumption below is standard~\cite{BeckBook2017},
and the assumption on the Hessian was used in \cite{jin2017escape} (for example, it is true if $f$ is quadratic). 

\begin{assumption}[Lipschitz Properties]\label{as::smooth_Lip}
$\nabla f$ is $\Lip$-Lipschitz continuous and $\nabla^2 f$ is $\rho$ Lipschitz continuous. We write $\H$ as shorthand for $\nabla^2 f(\x)$ when $\x$ is clear from context.
\end{assumption}

\begin{shortversion}
\begin{assumption}[Moderate Nonsmooth Term]
\label{as::small_lambda}
The magnitude of $\|\x\|_1$ term, denoted by $\lambda$, is small. \footnote{The quantification of $\lambda$ is given in the extended version of this paper.}
\end{assumption}
\end{shortversion}

\begin{longversion}
\begin{assumption}[Moderate Nonsmooth Term]
\label{as::small_lambda}
The magnitude of $\|\x\|_1$ term, which is denoted by $\lambda$, satisfies inequalities \eqref{eqn::lambda_bound} and \eqref{eqn::lambda_bound2}.
\end{assumption}
\end{longversion}



\begin{theorem}[Main]
\label{thm::main}
There exists an absolute constant $c_{\max}$ such that if $f(\cdot)$ satisfies \ref{as::smooth_Lip} and \ref{as::small_lambda},  then for any $\delta>0, \varepsilon \le \frac{\Lip^2}{\rho}, \Delta_\Phi \ge \Phi(\x_0) - \Phi^\star$, and constant $c \le c_{\max}$, with probability $1-\delta$, the output of $\text{PPD}(\x_0, \Lip, \rho, \varepsilon, c, \delta, \Delta_f)$ will be a $\varepsilon$-second order stationary point, and terminate in iterations:
\begin{equation*}
\mathcal{O}\left(\frac{\Lip(\Phi(\x_0) - \Phi^\star)}{\varepsilon^2}\ln^{4}\left(\frac{d\Lip\Delta_\Phi}{\varepsilon^2\delta}\right) \right)
\end{equation*}
\end{theorem}

\paragraph{Remark} Assuming $\varepsilon\le\frac{\Lip^2}{\rho}$ does not lead to loss of generality. Recall the second order condition is specified as $\lambda\big(\nabla^2 f(\x^\star)\big)_{\min} \ge -\sqrt{\rho\varepsilon}$, since when $\varepsilon\ge\frac{\Lip^2}{\rho}$, we always have $-\sqrt{\rho\varepsilon}\le -\Lip \le \lambda\big( \nabla^2f(\x^\star)  \big)_{\min} $, where the second inequality follows from the fact that the Lipschitz constant is the upper bound for $\lambda(\nabla^2 f(\x))$ in norm. Consequently, when $\varepsilon \ge \frac{\Lip^2}{\rho}$, every \ESSP is automatically a first order stationary point.

\begin{shortversion}
\subsection{Sketch of the proof of main theorem\footnote{An extended version of our paper with the full details of the proofs is available at \url{http://amath.colorado.edu/faculty/becker/PerturbedProximalDescent.pdf}.}}
We consider the objective function $\Phi = f+g = f(\x) + \lambda\|\x\|_1$. When the magnitude of the $\ell_1$ penalty term is small so that $f$ dominates the geometric landscape of the function $\Phi$, we expect that the characteristics of the objective function should not be too different from the without-$\ell_1$ penalty case. At a high-level, we follow the proof of \cite{jin2017escape}.

The key intuition is that when the iteration arrives in the vicinity of a saddle point, the volume of the trapping region surrounding the saddle point is small. As there is at least one direction for function value to continue decreasing \big(e.g., the eigenvector corresponding to $\lambda\big(\nabla^2f(\tilde{\x})\big)_{\min}\le -\gamma$\big), a random perturbation $\boldsymbol{\xi}$ added to the current iterate
will likely have a component in the escape direction.

We first show that when the iteration arrives in the vicinity of a saddle point $\tilde{\x}$, before achieving sufficient decrease in function value, which partially determines a time threshold $T$, the proximal descent iteration will remain bounded around the saddle point; i.e., $\|\u_t - \tilde{\x}\|\le \mathrm{const}$ for all $t< T$.

Introducing a perturbation will take the current iteration point $\u_0$ to $\w_0 = \u_0 + \boldsymbol{\xi}$. We track the development of these two iteration sequences $\{\u_t\}$ and $\{\w_t\}$ when proximal descent is applied. We show that when the magnitude of the nonsmooth $\ell_1$ term $\lambda$ is less than a certain constant $\Lambda$, these two iteration sequences will stay at least a fixed distance apart at every step; i.e., $\|\w_t - \u_t\|\ge \mathrm{const}$ for all $t<T$.

The central idea of proving the perturbed sequence will escape the saddle points is that after the perturbation is introduced, the projection of the iteration sequence in the escaping directions, i.e., on the subspace spanned by eigenvectors of negative eigenvalues of $\nabla^2 f(\tilde{\x})$, will gain more and more weight. 
To quantify this central idea, an key observation is that when magnitude of the $\ell_1$ penalty term is small, the proximal step will preserve the monotonicity relation between increasing weight of iterations on escape-beneficial subspaces and the iteration progress.

Combining the previous two results, we show that at least one of these two iteration sequences will attain sufficient decrease in function value within the given time threshold $T$, to be followed by the argument that the probability of the chosen perturbation not letting the perturbed iteration sequence to escape the saddle point is small.

The key issue in the final step of the proof is to ensure the returned result will be an \ESSP; in other words, we will show that whenever the current point is not an \ESSP, the algorithm cannot terminate, combined with the proof that the proposed algorithm \ref{algo::PPD} will terminate within finitely many steps. 

\end{shortversion}

\begin{longversion}

For the proof of the main theorem, we introduce some \underline{notation and units} for the simplicity of proof statement.

For matrices we use $\|\cdot\|$ to denote spectral norm. The operator $\mathcal{P}_\mathcal{S}(\cdot)$ denotes projection onto set $\mathcal{S}$. Define the local approximation of the smooth part of the objective function by
\begin{equation}
\tilde{f}_{\x}(\y)\defeq f(\x)+\nabla^T f(\x)(\y-\x)+\frac{1}{2}(\y-\z)^T\H(\y-\z) \label{eqn:f-approx}
\end{equation}

\paragraph{Units} With the conditional number of the Hessian matrix $\kappa \defeq \frac{\Lip}{\gamma}\ge 1$, we define the following units for the convenience of proof statement:
\begin{align*}
&\ufun \defeq \eta \Lip \frac{\gamma^3}{\rho^2}\cdot \ln^{-3}(\logterms),
&&\ugrad \defeq \sqrt{\eta \Lip} \frac{\gamma^2}{\rho} \cdot \ln^{-2}(\logterms)\\
&\uspace \defeq \sqrt{\eta\Lip}\frac{\gamma}{\rho} \cdot \ln^{-1}(\logterms),
&&\utime \defeq \frac{\ln (\logterms)}{\eta \gamma}
\end{align*}



\subsection{Lemma: Iterates remain bounded if stuck near a saddle point}

\begin{lemma}\label{lem:1st_seq}
For any constant $\ca \ge 3$, there exists absolute constant $c_{\max}$: for any $\delta\in (0, \frac{d\cn}{e}]$, let $f(\cdot), \tilde{\x}$ satisfies the condition in Lemma \ref{lem::main_lemma}, for any initial point $\u_0$ with $\norm{\u_0 - \tilde{\x}} \le 2\uspace/(\cn\cdot \ln(\logterms))$, define: 
\begin{equation*}
T = \min\left\{ ~\inf_t\left\{t \,|\, \tilde{f}_{\u_0}(\u_t) - f(\u_0) + g(\u_t) - g(\u_0)   \le -3\ufun \right\},  \ca\utime~\right\}
\end{equation*}
then, for any $\eta \le c_{\max}/\Lip$, we have for all $t<T$ that $\norm{\u_t - \tilde{\x}} \le 100( \uspace\cdot \ca )$.
\end{lemma}

\begin{proof}
We show if the function value did not decrease, then all the iteration updates must be constrained in a small ball. 
The proximal descent updates the solution as
\begin{align*}
\tilde{\u}_{t+1} &= \u_t - \nabla f(\u_t) = (I-\nabla f)(\u_t)\\
\u_{t+1} &= \mathrm{prox}_{\eta g}\big(\tilde{\u}_{t+1}\big) = \mathrm{prox}_{\eta g}\circ (I-\nabla f)(\u_t)
\end{align*}
Without losing of generality, set $\u_0 = 0$ to be the origin. For any $t\in\mathbb{N}$,
\begin{align*}
\|\u_{t} - \u_0\| &= \|\u_t - 0\| = \|\mathrm{prox}_{\eta g}(\tilde{\u}_{t}) - \mathrm{prox}_{\eta g}(0) \| \le \|\tilde{\u}_{t}-0\| = \|\tilde{\u}_t\|
\end{align*}
Jin et al. prove in \cite{jin2017escape} by induction that if $\|\u_t\|\le 100(\uspace\cdot\ca)$, then $\|\tilde{\u}_{t+1}\|\le 100(\uspace\cdot \ca)$. Consequently, $\|\u_{t+1}\|\le 100(\uspace\cdot \ca)$.\\

We point out that it is implicitly assumed that $\frac{2\uspace}{\kappa\cdot\ln(\logterms)} \ll \ca$, so that for all $t<T$, $\|\tilde{\x}\|\ll\|\u_t\|$, and the relation $\|\u_t - \tilde{\x}\|\le\|\u_t\|+\|\tilde{\x}\|\le 100(\uspace\cdot\ca)$ holds.
\end{proof}

\subsection{Preparation for Building Pillars}
\begin{lemma}[Existence of lower bound for the difference sequence $\{\v_t\}_{t=1}^{T}$]
For iteration sequences $\{\w_t\}$ and $\{\u_t\}$ defined in Lemma \ref{lem:2nd_seq}, define the difference sequence as
\[
\v_t = \w_t - \u_t
\]
There exists a positive lower bound for $\{\v_t\}$ when $t<\ca\utime$.
\label{lem::Existence_Lower_Bound}
\end{lemma}
\begin{proof}
To show that the lower bound for iteration difference $\{\v_t\}_{t=1}^{T}$ exists, we consider bounding the iteration sequence $\tilde{\v}_{t+1}$ first. Define the difference between the proximal of $l_1$ penalty term and its coimage as $\mathcal{D}_{ g}[\x] = \prox_{ g}[\x] - \x = \min\{\lambda \mathbbm{1} , |\x| \} \otimes \, \mathrm{sgn}(-\x)$, where $\otimes$ is Hadamard product and the minimum is taken elementwise. 
We notice that  $\|\mathcal{D}_{\eta\lambda\|\cdot\|_1}[\x]\| \le \eta\lambda\sqrt{d}$. Thus, $\|\w_k - \u_k\| = \|\tilde{\w}_k - \tilde{\v}_k - \lambda(\mathcal{D}_{\eta g}[\tilde{\w}_k] - \mathcal{D}_{\eta g}[\tilde{\u}_k])\| \ge \|\tilde{\w}_k - \tilde{\v}_k\| - 2\eta\lambda\sqrt{d}$.
\begin{align*}
    \|\tilde{\v}_{t+1}\| &= \|\tilde{\w}_{t+1} - \tilde{\u}_{t+1}\|\\
    &= \|(I-\eta\nabla f)\circ\prox_{\eta g}(\tilde{\w}_k) - (I-\eta\nabla f)\circ\prox_{\eta g}(\tilde{\u}_k)\|\\
    &= \|\w_k - \u_k - \eta(\nabla f(\w_k) - \nabla f(\u_k))\|\\
    &\ge \|\w_k-\u_k\| - \eta \Lip\|\w_k-\u_k\|= (1-\eta \Lip)\|\w_k-\u_k\|\\
    &\ge (1-\eta \Lip)(\|\tilde{\w}_k - \tilde{\u}_k\|-2\eta\lambda\sqrt{d}) = (1-\eta \Lip)(\|\tilde{\v}_k\|-2\eta\lambda\sqrt{d})\\
    &\ge (1-\eta \Lip)^t\|\tilde{\v}_1\| - 2\eta\lambda\sqrt{d}\sum_{i=1}^{t}(1-\eta \Lip)^i\\
    &= (1-\eta \Lip)^t\|\tilde{\v}_1\| - 2\lambda\sqrt{d}\frac{(1-\eta \Lip)\big(1-(1-\eta \Lip)^t\big)}{\Lip}
\end{align*}
As $\tilde{\v}_1 = (I-\eta\nabla f)\v_0 = (I-\eta\nabla f)\mu r \e_1 = \mu r(\e_1 - \eta\nabla^2 f(\boldsymbol{\xi})\theta \e_1) = \mu r (1 + \eta\gamma\theta)\e_1$, where $\theta\in(0,1)$, we have
\begin{equation}
    \|\tilde{\v}_{t+1}\|\ge (1-\eta \Lip)^t\mu r(1+\eta\gamma\theta) - 2\lambda\sqrt{d}\frac{(1-\eta \Lip)(1-(1-\eta \Lip)^t)}{\eta \Lip}
\end{equation}

To compare $\|\v_t\|$ and $\|\tilde{\v}_t\|$, \begin{equation}
    \|\v_{t+1}\|\ge\|\tilde{\v}_{t+1}\| - 2\eta\lambda\sqrt{d} \ge (1-\eta \Lip)^t\mu r(1+\eta\gamma\theta) - 2\lambda\sqrt{d}\frac{(1-\eta \Lip)(1-(1-\eta \Lip)^t)+\eta \Lip}{\Lip}
\end{equation}
Therefore, as long as 
\begin{equation}
\label{eqn::lambda_bound}
\lambda < \frac{(1-\eta \Lip)^{\ca\utime}\mu\frac{1}{ \kappa(\ln \logterms)^2}\sqrt{\eta} \Lip^{\frac{3}{2}} \frac{\gamma}{\rho}(1+\eta\gamma\theta) }{2\sqrt{d}[(1-\eta \Lip)(1-(1-\eta \Lip)^{\ca\utime})+\eta \Lip]}
\end{equation}
the difference sequence $\{\|\v_t\|\}$ has a positive lower bound on its norm.
\end{proof}

\begin{lemma}[Preservation of subspace projection monotonicity after prox of $l_1$ in rotated coordinate with small $\lambda$] 
\label{lem::Monotonicity}

Denote the subspace of $\R^n$ spanned by $\{\e_1\}$ as $\mathbb{E}$, while the complement subspace spanned by $\{\e_2,\cdots,\e_n\}$ as $\mathbb{E}^\perp$. For a given vector $\x$ chosen from a lower bounded set $\mathcal{X}$, i.e. $\forall \, \x \in \mathcal{X}$, $\|\x\|\ge C$ for some constant $C>0$, assume $ \|\proj_{\mathbb{E}^\perp} \x\| \le K \|\proj_{\mathbb{E}}\x\| $, where $0 < K \le 1$ is a constant.
If the parameter $\lambda$ for the $l_1$ penalty term is small enough, then 
\[
\|\proj_{\mathbb{E}^\perp}\prox_{\eta g}(\x)\| \le K\|\proj_{\mathbb{E}}\prox_{\eta g}(\x)\|
\]
\end{lemma}
\begin{proof}



We want to find a constraint on $\lambda$ such that when $\lambda$ is small enough, if the projection in the original coordinate demonstrates the monotonicity relation $\|\proj_{\mathbb{E}}\x\|\le\|\proj_{\mathbb{E}^\perp} \x\|$, this monotonicity relation will be preserved after proximal operator of $l_1$ is applied on the input vector.\\

Naturally there exists a normal vector, denoted as $\hat{\n}_{\text{boundary}}\equiv\hat{\n}$, for the boundary hyperplane on which $\|\proj_{\mathbb{E}} \x\| = K\|\proj_{\mathbb{E}^\perp} \x\|$. By moving along $\hat{\n}$, a point approaches the boundary most efficiently. Any vector inside the hyperplane is perpendicular to $\hat{\n}$, which we denote as $\hat{\n}^\perp$.

Define \begin{equation}
\label{eqn::vmove_vec_defn}
\hat{\v}_{\textrm{move}}(\x) = \begin{cases}
-\eta\lambda \cdot\mathrm{sgn}(x_i) & \textrm{ if } |x_i| \ge \eta\lambda \\
-x_i & \text{ if } |x_i| < \eta\lambda
\end{cases} = \min\{|x|,\eta\lambda\mathbbm{1}\}\otimes \mathrm{sgn}(-\x)
\end{equation} where $\otimes$ is the Hadamard product, and the minimum is taken elementwise. Because $\prox_{\eta g}(\x) = \x + \hat{\v}_{\textrm{move}} $, a sufficient condition to be imposed on $\lambda$ to guarantee the preservation of projection monotonicity $\|\proj_{\mathbb{E}^\perp}\prox_{\eta g}(\x)\| \le K\|\proj_{\mathbb{E}}\prox_{\eta g}(\x)\|$ is that 
\[
\lambda < \bigg\| \frac{ \mathrm{Proj}_{\n}\x }{\hat{\v}_{\mathrm{move}} \cdot \hat{\n}} \bigg\| = \bigg\| \frac{ \x \cdot \hat{\n} }{\hat{\v}_{\mathrm{move}} \cdot \hat{\n}} \bigg\| \le \frac{\|\x\|}{\|\hat{\v}_{\mathrm{move}} \cdot \hat{\n} \|}
\]
which means the moving distance caused by applying the $l_1$ proximal operator (soft shrinkage) projected on the direction of $\hat{\n}$ is less that the distance between $\x$ to the boundary hyperplane, hence rendering the vector stay on the same side of the boundary after moving.\\

Therefore, as long as 
\begin{equation}
\label{eqn::lambda_bound2}
\lambda < \frac{C}{\|\hat{\v}_{\mathrm{move} } \cdot \hat{\n} \|}
\end{equation} the monotonicity of projection onto subspaces can be preserved.

\end{proof}

\paragraph{Remark 1 for Lemma \ref{lem::Monotonicity}} As an examples in $\R^2$, set $K=1$, we visualise the shift caused by proximal operator and the boundary of projection-monotonicity preserving region. Assume $\e_{1,2}$ are orthonormal basis of Cartesian coordinate in the standard position. The directional vector for region division boundary is $\dis \hat{\e}_{\mathrm{boundary}} = \hat{\n}^\perp = \frac{\pm\hat{\e}_1\pm\hat{\e}_2}{\sqrt{2}} $, and $\hat{\e}_{\mathrm{boundary}}^\perp = \hat{\n}$ is the corresponding perpendicular directional vector. For $l_1$ norm, $\hat{\v}_{\mathrm{move}}$ is $(\pm 1, \pm 1)$.
\paragraph{Remark 2 for Lemma \ref{lem::Monotonicity}} We point out that the upper bound for the parameter $\lambda$ is related to the alignment of the eigenspace of $\H$. If the eigenspace of $\H$ is aligned with canonical orthonormal basis of $\R^d$, then $\lambda\in(0,\infty)$. The most stringent restriction on the upper bound of $\lambda$ applies when $\hat{\v}_{\textrm{move}}$ is parallel to $\hat{\n}$.






\subsection{Lemma: Perturbed iterates will escape the saddle point}

\begin{lemma}\label{lem:2nd_seq}
There exists absolute constant $c_{\max}, \ca$ such that: for any $\delta\in (0, \frac{d\cn}{e}]$, let $f(\cdot), \tilde{\x}$ satisfies the condition in Lemma \ref{lem::main_lemma}, and
sequences $\{\u_t\}, \{\w_t\}$ satisfy the conditions in Lemma \ref{lem::main_lemma}, define:
\begin{equation*}
T = \min\left\{ ~\inf_t\left\{t| \tilde{f}_{\w_0}(\w_t) + g(\w_t) - f(\w_0) - g(\w_0)  \le -3\ufun \right\},  \ca\utime~\right\}
\end{equation*}
then, for any $\eta \le c_{\max} / \Lip$, if $\norm{\u_t - \tilde{\x}} \le 100 (\uspace\cdot \ca )$ for all $t<T$, we will have $T < \ca \utime$.
\end{lemma}

\begin{proof}
We show that if the iterate sequence before time $T$ starting from $\u_0$ does not provide sufficient function value decrease, the other iterate sequence, which starts from $\w_0$, will be able to achieve the function value decrease purpose. Ultimately, we will prove $T<\ca\utime$.
We establish the inequality about $T$ by considering the difference between $\w_t$ and $\u_t$. Define $\v_t = \w_t - \u_t$. The assumption of the lemma \ref{lem:2nd_seq}, $\v_0 = \mu[\uspace/(\kappa\cdot\ln(\logterms))]\e_1$, $\mu\in[\delta/(2\sqrt{d}),1].$\\

We bound $\|\v_t\|$ from both sides for all $t<T$ to obtain an inequality about $T$. 

Recall that the proximal descent updates the solution as
\begin{align*}
\tilde{\u}_{t+1} &= \u_t - \nabla f(\u_t) = (I-\eta\nabla f)(\u_t)\\
\u_{t+1} &= \mathrm{prox}_{\eta g}\big(\tilde{\u}_{t+1}\big) = \mathrm{prox}_{\eta g}\circ (I-\eta\nabla f)(\u_t)
\end{align*}
Simple algebraic computation gives 
\begin{equation}
\label{dyn::v}
\tilde{\v}_{t+1} = (I-\eta\H-\eta\Delta_t')\v_t
\end{equation}
where $\Delta'_t = \int_0^1 \nabla^2f(\u_t+\theta\v_t)\,\mathrm{d} \theta - \H$, and $\tilde{\v}_{t} = \tilde{\w}_t - \tilde{\u}_t$. \\

Consider $\|\tilde{\u}_t\|$ and $\|\tilde{\w}_t\|$. Because $\v_0=\tilde{\v}_0$, we have $\|\tilde{\w}_0-\tilde{\x}\|\le \|\tilde{\u}_0-\tilde{\x}\|+\|\tilde{\v}_0\|\le 2 \uspace/(\kappa\cdot\ln(\logterms))$. With same logic in the proof for lemma \ref{lem:1st_seq}, we see $\|\tilde{\u}_t\|\le 100(\uspace\cdot\ca)$, and $\|\tilde{\w}_t\|\le 100(\uspace\cdot\ca)$. (Same relation hold for $\|\u_t\|$ and $\|\w_t\|$ respectively.) As a result, $\|\tilde{\v}_t\| \le \|\tilde{\w}_t\| + \|\tilde{\u}_t\| \le 200(\uspace\cdot\ca)$ for all $t<T$. Also, \begin{equation}
\label{eqn::v_upper_bound}
    \|\v_t\| \le 200(\uspace\cdot\ca)
\end{equation} 

Equation \eqref{eqn::v_upper_bound} and Hessian Lipschitz gives for $t<T$,  $\|\Delta'_t\|\le \rho(\|\u_t\|+\|\v_t\|+\|\tilde{\x}\|) \le  \rho\uspace(300\ca+1) = \frac{\zeta}{\eta}$, where $\zeta = \eta\rho\uspace(300\ca+1)$. \\


Denote $\psi_t$ be the norm of $\v_t$ projected onto $\e_1$ direction ($\S$), and $\varphi_t$ be the norm of $\v_t$ projected onto the remaining subspace ($\S^c$), while $\tilde{\psi}_t$ be the norm of $\tilde{\v}_t$ projected onto $\S$, and $\tilde{\varphi}_t$ be the norm of $\tilde{\v}_t$ projected onto $\S^c$. 


Equation \eqref{dyn::v} gives
\begin{align}
\tilde{\psi}_{t+1} &\ge (1+\gamma\eta) \psi_t - \zeta\sqrt{\psi_t^2+\varphi_t^2}\\
\tilde{\varphi}_{t+1} &\le (1+\gamma\eta)\varphi_t + \zeta\sqrt{\psi_t^2+\varphi_t^2}
\end{align}

To obtain the lower bound of $\|\v_t\|$, we prove the following relation as preparation:
\begin{equation}
\label{dyn::bridge}
\textrm{for all }t<T, \quad \varphi_t\le 4\zeta t\cdot  \psi_t
\end{equation}
By hypothesis of lemma \ref{lem:2nd_seq}, we know $\varphi_0 = 0$, thus the base case of induction holds. Assume equation \eqref{dyn::bridge} is true for $\tau \le t$, for $t+1\le T$, we have
\begin{align}
\label{eqn::dyn_bridge}
\tilde{\varphi}_{t+1} &\le 4\zeta t(1+\gamma\eta) \psi_t + \zeta \sqrt{\psi_t^2+\varphi_t^2} \nonumber\\
 4\zeta(t+1)\bigg[ (1+\gamma\eta)\psi_t - \zeta \sqrt{\psi_t^2+\varphi_t^2} \bigg] &\le 4\zeta(t+1)\tilde{\psi}_{t+1}
\end{align}
By choosing $\sqrt{c_{\max}}\le \frac{1}{300\ca+1}\min\{\frac{1}{2\sqrt{2}},\frac{1}{4\ca}\}$, and $\eta\le\frac{c_{\max}}{\Lip}$, we have $4\zeta(t+1)\le 4\zeta T\le 4\eta\rho\uspace(300\ca+1)\ca\utime = 4\sqrt{\eta\Lip}(300\ca+1)\ca\le 1$. This gives $4(1+\gamma\eta)\psi_t\ge 4\psi_t\ge (1+1)\sqrt{2\psi_t^2}\ge(1+4\zeta(t+1))\sqrt{\psi_t^2+\varphi_t^2}$.
i.e.
\begin{equation}
\label{dyn::bridge_sufficient_cond}
(1+4\zeta(t+1))\sqrt{\psi_t^2+\varphi_t^2} \le 4\psi_t 
\end{equation}
Connecting two parts of equation \eqref{eqn::dyn_bridge}, we obtain 
\begin{equation}
\label{dyn::dynamic_middle_step}
    \tilde{\varphi}_{t+1} \le 4\zeta(t+1)\tilde{\psi}_{t+1}
\end{equation}

Now we switch our focus to the eigenspace of Hessian $\H$. Assume the orthonormal basis for the eigensapce of $\H$ is $\{\e_1,\e_2,\cdots,\e_d\}$. The order of dimension aligns with the increasing order of the corresponding eigenvalues. This coordinate transformation does not lead to loss of generality, as it is unitary.\\
By lemma \ref{lem::Existence_Lower_Bound}, we know the iteration difference sequence $\v_t$ has a positive lower bound in terms of 2-norm. Therefore, by lemma \ref{lem::Monotonicity}, with the virtue of equation \eqref{dyn::dynamic_middle_step} $\sqrt{\sum_{i=2}^d (\e_i^T\tilde{\v}_{t+1})^2}\le 4\zeta(t+1)\|\e_1^T\tilde{\v}_{t+1}\|$, we still have the projection monotonicity on the subspace of eigenspace of $\H$, i.e. \[\varphi_{t+1} = \sqrt{\sum_{i=2}^d (\e_i^T \prox_g(\tilde{\v}_{t+1}))^2} \le 4\zeta(t+1)\|\e_1^T\prox_g(\tilde{\v}_{t+1})\| = 4\zeta(t+1)\psi_{t+1}\] 
Until here we finish the induction.





Recall that $ 4\zeta(t+1)\le 1 $, we thus have $\varphi_t\le 4\zeta t\psi_t \le \psi_t$, which gives
\begin{equation}
\label{eqn::recursion}
\psi_{t+1} \ge (1+\gamma\eta)\psi_t - \sqrt{2}\zeta\psi_t \ge \big( 1+\frac{\gamma\eta}{2}\big) \psi_t 
\end{equation}
where the last inequality follows from $\zeta=\eta\rho\uspace(300\ca + 1) \le \sqrt{c_{\max}}
(300\ca + 1)\gamma\eta\cdot\ln^{-1}(\logterms)\le \frac{\gamma\eta}{2\sqrt{2}}$.

Finally, combining \eqref{eqn::v_upper_bound} and \eqref{eqn::recursion}, we have for all $t<T$:
\begin{align*}
200(\uspace\cdot\ca) &\ge \|\v_t\| \ge \psi_t \ge (1+\frac{\gamma\eta}{2})^t\psi_0 = (1+\frac{\gamma\eta}{2})^tc_0\frac{\uspace}{\kappa}\ln^{-1}\bigg(\logterms\bigg)\\
&\ge (1+\frac{\gamma\eta}{2})^t \frac{\delta}{2\sqrt{d}}\frac{\uspace}{\kappa}\ln^{-1}\bigg(\logterms\bigg)
\end{align*}
This implies
\[
T<\frac{1}{2}\frac{\ln[400\frac{\kappa\sqrt{d}}{\delta}\cdot\ca\ln(\logterms)]}{\ln(1+\frac{\gamma\eta}{2})} \le \frac{\ln[400\frac{\kappa\sqrt{d}}{\delta}\cdot\ca\ln(\logterms)]}{\gamma\eta} \le (2+\ln(400\ca))\utime
\]
The last inequality is due to $\delta\in(0,\frac{d\kappa}{e}]$, we have $\ln(\logterms)\ge 1$. By choosing the constant $\ca$ to be large enough to satisfy $2+\ln(400\ca)\le \ca$, we will have $T<\ca\utime$, which finishes the proof.




\end{proof}

\subsection{Combining Previous Results}
\begin{lemma}\label{lem:one_in_two}
There exists a universal constant $c_{\max}$, for any $\delta\in (0, \frac{d\cn}{e}]$, let $f(\cdot), \tilde{\x}$ satisfies the conditions in Lemma \ref{lem::main_lemma}, and without loss of generality let $\e_1$ be the minimum eigenvector of $\hess f(\tilde{\x})$. Consider two gradient descent sequences $\{\u_t\}, \{\w_t\}$ with initial points $\u_0, \w_0$ satisfying: (denote radius $r = \uspace/(\cn\cdot \ln(\logterms))$)
\begin{equation*}
\norm{\u_0 - \tilde{\x}} \le r, \quad \w_0 = \u_0 +\mu \cdot r \cdot \e_1, \quad\mu \in [\delta/(2\sqrt{d}), 1]
\end{equation*}
Then, for any stepsize $\eta \le c_{\max} / \Lip$, and any $T \ge \frac{1}{c_{\max}}\utime$, we have:
\begin{equation*}
\min\{f(\u_{T}) + g(\u_T)  - f(\u_0) - g(\u_0), f(\w_{T}) + g(\w_T) - f(\w_0) - g(\w_0)\} \le -2.7\ufun
\end{equation*}
\end{lemma}
\begin{proof}
Without losing generality, let $\tilde{\x} = 0$ be the origin. Let $(c^{(2)}_{\max}, \ca)$ be the absolute constant so that Lemma \ref{lem:2nd_seq} holds, also let $c^{(1)}_{\max}$ be the absolute constant to make Lemma \ref{lem:1st_seq} holds based on our current choice of $\ca$.
We choose $c_{\max} \le \min\{c^{(1)}_{\max}, c^{(2)}_{\max}\}$ so that our learning rate $\eta \le c_{\max}/\Lip$ is small enough which make both Lemma \ref{lem:1st_seq} and Lemma \ref{lem:2nd_seq} hold. Let $T^\star \defeq \ca\utime$ and define:
\begin{equation*}
\Ts = \inf_t\left\{t| \tilde{f}_{\u_0}(\u_t) + g(\u_t) - f(\u_0) - g(\u_0)  \le -3\ufun \right\}
\end{equation*}
Let's consider following two cases:

\paragraph{Case $\Ts \le T^\star$:} In this case, by Lemma \ref{lem:1st_seq}, we know $\norm{\u_{\Ts-1}} \le O(\uspace)$, and therefore
\begin{align*}
\norm{\u_{\Ts}} \le &\norm{\u_{\Ts-1}} + \eta \norm{\grad f(\u_{\Ts-1})}
\le \norm{\u_{\Ts-1}} + \eta \norm{\grad f(\tilde{\x})} + \eta \Lip \norm{\u_{\Ts-1}} 
\le O(\uspace)
\end{align*}
By choosing $c_{\max}$ small enough and $\eta \le c_{\max} /\Lip$, this gives:
\begin{align*}
&f(\u_{\Ts}) + g(\u_{\Ts}) - f(\u_0) - g(\u_0)\\ &\le \grad f(\u_0)\trans (\u_{\Ts}-\u_0) + \frac{1}{2}(\u_{\Ts}-\u_0)\trans \hess f(\u_0) (\u_{\Ts}-\u_0)
+ \frac{\rho}{6} \norm{\u_{\Ts}-\u_0}^3 + g(\u_{\Ts}) - g(\u_0) \\
&\le \tilde{f}_{\u_0}(\u_{\Ts}) - f(\u_0) + g(\u_{\Ts}) - g(\u_0) + \frac{\rho}{2}\norm{\u_0 - \tilde{\x}}\norm{\u_{\Ts}-\u_0}^2+ \frac{\rho}{6} \norm{\u_{\Ts}-\u_0}^3 \\
&\le -3\ufun + O(\rho \uspace^3) = -3\ufun + O(\sqrt{\eta\Lip}\cdot\ufun) \le -2.7\ufun
\end{align*}
The first and second inequality exploit Hessian Lipschitz property of smooth function $f$, and $\|\u_0-\tilde{\x}\|\le O(\uspace)$, $\|\u_{T'} - \u_0\|\le O(\uspace)$. By choose $c_{\max} \le \min \{1, \frac{1}{\ca}\}$. We know $\eta < \frac{1}{\Lip}$, by \textit{sufficient decrease lemma} for proximal descent, we know each proximal descent iteration decreases function value. Therefore, for any $T\ge \frac{1}{c_{\max}}\utime \ge \ca\utime = T^\star \ge \Ts$, we have:
\begin{equation*}
\Phi(\u_T) - \Phi(\u_0) \le \Phi(\u_{T^\star})  - \Phi(\u_0) \le \Phi(\u_{\Ts}) - \Phi(\u_0) \le -2.7\ufun
\end{equation*}

\paragraph{Case $\Ts > T^\star$:} In this case, by Lemma \ref{lem:1st_seq}, we know $\norm{\u_t}\le O(\uspace )$ for all $t\le T^\star$. Define
\begin{equation*}
\Tt = \inf_t\left\{t| \tilde{f}_{\w_0}(\w_t) + g(\w_t) - f(\w_0) - g(\w_0)  \le -3\ufun \right\}
\end{equation*}
By Lemma \ref{lem:2nd_seq}, we immediately have $\Tt \le T^\star$. Apply same argument as in the case $T'\le T^\star$, we have for all $T\ge \frac{1}{c_{\max}}\utime$ that $f(\w_T) + g(\w_T) - f(\w_0) - g(\w_0) \le f(\w_{T^\star}) + g(\w_{T^\star})  - f(\w_0) - g(\w_0) \le -2.7\ufun$.
\end{proof}

\subsection{Main Lemma}

\begin{lemma}[Main Lemma]\label{lem::main_lemma}
There exists universal constant $c_{\max}$, for $f(\cdot)$ satisfies \ref{as::smooth_Lip}, for any $\delta\in (0, \frac{d\cn}{e}]$, suppose we start with point $\tilde{\x}$ satisfying following conditions:
\begin{equation*}
\|G(\tilde{\x})\|=\bigg\|\Lip(\tilde{\x}-\mathrm{prox}_{\frac{1}{\Lip}g}\bigg(\tilde{\x}-\frac{1}{L}\nabla f(\tilde{\x})\bigg)\bigg\| \le \ugrad \quad  \text{~and~} \quad  \lambda_{\min}(\hess f(\tilde{\x})) \le -\gamma
\end{equation*}
Let $\x_0 = \tilde{\x} + \boldsymbol{\xi}$ where $\boldsymbol{\xi}$ come from the uniform distribution over ball with radius $\uspace /(\cn\cdot\ln(\logterms))$, 
and let $\x_t$ be the iterates of gradient descent from $\x_0$. Then, when stepsize 
$\eta \le c_{\max} / \Lip$, with at least probability $1-\delta$, we have following for any $T \ge \frac{1}{c_{\max}}\utime$:
\begin{equation*}
f(\x_T) + g(\x_T) - f(\tilde{\x}) - g(\tilde{\x}) \le -\ufun
\end{equation*}
\end{lemma}

\begin{proof}
Denote $T_\frac{l}{\Lip}(\x) = \prox_{\frac{1}{\Lip}g} \big[ \x-\frac{1}{\Lip}\nabla f(\x) \big]$. The fisrt order stationary condition is equivalent to $\|\tilde{\x}-T_\frac{1}{\Lip}(\tilde{\x})\| = \|\nabla f(\tilde{\x}) + \partial g\big(T_\frac{1}{\Lip}(\tilde{\x})\big)\|\le\ugrad$, where $\partial g$ is the subgradient of the function $g$.

As $g(\x) = \lambda\|\x\|_1$ has Lipschitz constant $\lambda$, we have 
\[
f(\x_0) + g(\x_0) \le f(\tilde{\x}) + \<\nabla f(\tilde{\x}),\boldsymbol{\xi}\> + \frac{\Lip}{2}\|\boldsymbol{\xi}\|^2 + g(\tilde{\x}) + \< \partial g(\tilde{\x}),\boldsymbol{\xi}\> + \frac{\lambda}{2}\|\boldsymbol{\xi}\|^2
\]
Notice
\begin{align*}
\|\nabla f(\tilde{\x}) + \partial g(\tilde{\x})\| &= \|\nabla f(\tilde{\x}) + \partial g(T_\frac{l}{\Lip}(\x)) - \big(\partial g(T_\frac{l}{\Lip}(\x)) - \partial g(\tilde{\x})\big)\|\\
&\le \ugrad + \lambda \ugrad
\end{align*}
By adding perturbation, in worst case we increase function value by:
\begin{align*}
f(\x_0) - f(\tilde{\x}) + g(\x_0) - g(\tilde{\x}) & \le  \|\grad f(\tilde{\x})+\partial g(\tilde{\x})\|\|\xi\| +  \frac{\Lip + \lambda}{2} \norm{\xi}^2 \\
&\le
(1+\lambda)\ugrad(\frac{\uspace}{\cn \cdot \ln(\logterms)}) + \frac{1}{2}(\Lip+\lambda)(\frac{\uspace}{\cn \cdot \ln(\logterms)})^2 \\
&\le (\frac{3}{2}+\frac{1}{5})\ufun
\end{align*}
where the last inequality follows from the fact that $\lambda \ll \min\{1,l\}$ per equation \eqref{eqn::lambda_bound}.

On the other hand, let radius $r = \frac{\uspace}{\cn \cdot \ln(\logterms)}$. We know $\x_0$ come froms uniform distribution over $\ball_{\tilde{\x}}(r)$. Let $\cXs \subset \ball_{\tilde{\x}}(r)$ denote the set of bad starting points so that if $\x_0 \in \cXs$, then $\Phi(\x_T) - \Phi(\x_0) > -2.7\ufun$ (thus stuck at a saddle point); otherwise if $\x_0 \in B_{\tilde{\x}}(r) - \cXs$, we have $\Phi(\x_T) - \Phi(\x_0) \le -2.7\ufun$.

By applying Lemma \ref{lem:one_in_two}, we know for any $\x_0\in \cXs$, it is guaranteed that $(\x_0 \pm \mu r \e_1) \not \in \cXs $ where $\mu \in [\frac{\delta}{2\sqrt{d}}, 1]$. Denote $I_{\cXs}(\cdot)$ be the indicator function of being inside set $\cXs$; and vector $\x = (x^{(1)}, \x^{(-1)})$, where $x^{(1)}$ is the component along $\e_1$ direction, and $\x^{(-1)}$ is the remaining $d-1$ dimensional vector. Recall $\ball^{(d)}(r)$ be $d$-dimensional ball with radius $r$;  By calculus, this gives an upper bound on the volumn of $\cXs$:
\begin{align*}
\text{Vol}(\cXs) =& \int_{\ball^{(d)}_{\tilde{\x}}(r)}  \mathrm{d}\x \cdot I_{\cXs}(\x)\\
= & \int_{\ball^{(d-1)}_{\tilde{\x}}(r)} \mathrm{d} \x^{(-1)} \int_{\tilde{x}^{(1)} - \sqrt{r^2 - \norm{\tilde{\x}^{(-1)} - \x^{(-1)}}^2}}^{\tilde{x}^{(1)} + \sqrt{r^2 - \norm{\tilde{\x}^{(-1)} - \x^{(-1)}}^2}} \mathrm{d} x^{(1)}  \cdot  I_{\cXs}(\x)\\
\le & \int_{\ball^{(d-1)}_{\tilde{\x}}(r)} \mathrm{d} \x^{(-1)} \cdot\left(2\cdot \frac{\delta}{2\sqrt{d}}r \right) = \text{Vol}(\ball_0^{(d-1)}(r))\times \frac{\delta r}{\sqrt{d}}
\end{align*}
Then, we immediately have the ratio:
\begin{align*}
\frac{\text{Vol}(\cXs)}{\text{Vol}(\ball^{(d)}_{\tilde{\x}}(r))}
\le \frac{\frac{\delta r}{\sqrt{d}} \times \text{Vol}(\ball^{(d-1)}_0(r))}{\text{Vol} (\ball^{(d)}_0(r))}
= \frac{\delta}{\sqrt{\pi d}}\frac{\Gamma(\frac{d}{2}+1)}{\Gamma(\frac{d}{2}+\frac{1}{2})}
\le \frac{\delta}{\sqrt{\pi d}} \cdot \sqrt{\frac{d}{2}+\frac{1}{2}} \le \delta
\end{align*}
The second last inequality is by the property of Gamma function that $\frac{\Gamma(x+1)}{\Gamma(x+1/2)}<\sqrt{x+\frac{1}{2}}$ as long as $x\ge 0$.
Therefore, with at least probability $1-\delta$, $\x_0 \not \in \cXs$. In this case, we have:
\begin{align*}
\Phi(\x_T) - \Phi(\tilde{\x}) &= \Phi(\x_T)  - \Phi(\x_0) +  \Phi(\x_0) - \Phi(\tilde{\x}) \\
&\le  -2.7\ufun + 1.7\ufun \le -\ufun
\end{align*}
which finishes the proof.

\end{proof}

\subsection{Main Theorem, and its Proof}
\begin{lemma}[Sufficient Decrease Lemma for Proximal Descent, \cite{BeckBook2017}]
\label{lem::sufficient_decrease}
Assume the function $f$ is real-valued and lower semi-continuous. Then for any $L\in(\frac{\Lip}{2},\infty)$ where $\eta = \frac{1}{L}$, we have
$
\Phi(\x_t) - \Phi(\x_{t+1})\ge \frac{L-\frac{\Lip}{2}}{L^2}\|G_{\frac{1}{L}}(\x_t)\|.
$
\end{lemma}


\subsubsection{Proof of the Main Theorem}


\begin{proof} 
Denote $\tilde{c}_{\max}$ to be the absolute constant allowed in lemma \ref{lem::main_lemma} when it is given following parameters $\eta = \frac{c}{\Lip}$, $\gamma=\sqrt{\rho\varepsilon}$, and $\delta = \frac{d\Lip}{\sqrt{\rho\varepsilon}}\mathrm{e}^{-\chi}$.
In this theorem, we let $c_{\max} = \min\{\tilde{c}_{\max}, 1/2\}$, and choose any constant $c \le c_{\max}$.


In this proof, we will actually achieve some point satisfying following condition:
\begin{equation} \label{eq:tighter_cond}
 \norm{G(\x)} \le g_{\text{thres}} \equiv \frac{\sqrt{c}}{\chi^2} \cdot \varepsilon, \qquad\qquad \lambda_{\min}(\hess f(\x)) \ge - \sqrt{\rho \varepsilon}
\end{equation}
Since $c\le1$, $\chi\ge 1$, we have $\frac{\sqrt{c}}{\chi^2} \le 1$, which implies any $\x$ satisfy Eq.\eqref{eq:tighter_cond} is also a \ESSP.

Starting from $\x_0$, we know if $\x_0$ does not satisfy Eq.\eqref{eq:tighter_cond}, there are only two possibilities:
\begin{enumerate}
\item $\norm{G(\x_0)} > g_{\text{thres}}$: In this case, Algorithm \ref{algo::PPD} will not add perturbation. By lemma \ref{lem::sufficient_decrease}:
\begin{equation*}
\Phi(\x_{1}) - \Phi(\x_0) \le  -\frac{\eta}{2} \cdot g_{\text{thres}}^2 = -\frac{c^2}{2\chi^4}\cdot\frac{\varepsilon^2}{\Lip}
\end{equation*}

\item $\norm{G(\x_0)} \le g_{\text{thres}}$:
In this case, Algorithm \ref{algo::PPD} will add a perturbation of radius $r$, and will perform proximal gradient descent (without perturbations) for the next $t_{\text{thres}}$ steps. Algorithm \ref{algo::PPD} will then check termination condition. If the condition is not met, we must have:
\begin{equation*}
\Phi(\x_{t_{\text{thres}}}) - \Phi(\x_0) \le -\Phi_{\text{thres}} = -\frac{c}{\chi^3}\cdot\sqrt{\frac{\varepsilon^3}{\rho}}
\end{equation*}
This means on average every step decreases the function value by
\begin{equation*}
\frac{\Phi(\x_{t_{\text{thres}}}) - \Phi(\x_0)}{t_{\text{thres}}} \le -\frac{c^3}{\chi^4}\cdot\frac{\varepsilon^2}{\Lip}
\end{equation*}
\end{enumerate}
In case 1, we can repeat this argument for $t = 1$ and in case 2, we can repeat this argument for $t=t_{\text{thres}}$.
Hence, we can conclude as long as algorithm \ref{algo::PPD} has not terminated yet, on average, every step decrease function value by at least $\frac{c^3}{\chi^4}\cdot\frac{\varepsilon^2}{\Lip}$. However, we clearly can not decrease function value by more than $\Phi(\x_0) - \Phi^\star$, where $\Phi^\star$ is the function value of global minima. This means algorithm \ref{algo::PPD} must terminate within the following number of iterations:
\begin{equation*}
\frac{\Phi(\x_0) - \Phi^\star}{\frac{c^3}{\chi^4}\cdot\frac{\varepsilon^2}{\Lip}}
= \frac{\chi^4}{c^3}\cdot \frac{\Lip(\Phi(\x_0) - \Phi^\star)}{\varepsilon^2} = O\left(\frac{\Lip(\Phi(\x_0) - \Phi^\star)}{\varepsilon^2}\ln^{4}\left(\frac{d\Lip\Delta_\Phi}{\varepsilon^2\delta}\right) \right)
\end{equation*}


Finally, we would like to ensure when Algorithm \ref{algo::PPD} terminates, the point it finds is actually an \ESSP. The algorithm can only terminate when the gradient mapping is small, and the function value does not decrease after a perturbation and $t_{\text{thres}}$ iterations. We shall show every time when we add perturbation to iterate $\tilde{\x}_t$, if $\lambda_{\min}(\hess f(\tilde{\x}_t)) < - \sqrt{\rho \varepsilon}$, then
we will have $\Phi(\x_{t+t_{\text{thres}}}) - \Phi(\tilde{\x}_t) \le -\Phi_{\text{thres}}$. Thus, whenever the current point is not an \ESSP, the algorithm cannot terminate.

According to Algorithm \ref{algo::PPD}, we immediately know $\norm{G(\tilde{\x}_t)} \le g_{\text{thres}}$ (otherwise we will not add perturbation at time $t$). By lemma \ref{lem::main_lemma}, we know this event happens with probability at least $1-\frac{d\Lip}{\sqrt{\rho\varepsilon}}e^{-\chi}$ each time. On the other hand, during one entire run of Algorithm \ref{algo::PPD}, the number of times we add perturbations is at most:
\begin{equation*}
\frac{1}{t_{\text{thres}}} \cdot \frac{\chi^4}{c^3}\cdot \frac{\Lip(\Phi(\x_0) - \Phi^\star)}{\varepsilon^2}
=\frac{\chi^3}{c}\frac{\sqrt{\rho\varepsilon}(\Phi(\x_0) - \Phi^\star)}{\varepsilon^2}
\end{equation*}

By the union bound, for all these perturbations, with high probability lemma~\ref{lem::main_lemma} is satisfied. As a result Algorithm \ref{algo::PPD} works correctly. The probability of that is at least
\begin{equation*}
1- \frac{d\Lip}{\sqrt{\rho\varepsilon}}e^{-\chi} \cdot \frac{\chi^3}{c}\frac{\sqrt{\rho\varepsilon}(\Phi(\x_0) - \Phi^\star)}{\varepsilon^2}
= 1 -   \frac{\chi^3e^{-\chi}}{c}\cdot  \frac{d\Lip(\Phi(\x_0) - \Phi^\star)}{\varepsilon^2}
\end{equation*}

Recall our choice of $\chi = 3\max\{\ln(\frac{d\Lip\Delta_f}{c\varepsilon^2\delta}), 4\}$. Since $\chi \ge 12$, we have $\chi^3e^{-\chi} \le e^{-\chi/3}$, this gives:
\begin{equation*}
\frac{\chi^3e^{-\chi}}{c}\cdot  \frac{d\Lip(\Phi(\x_0) - \Phi^\star)}{\varepsilon^2}
\le e^{-\chi/3}  \frac{d\Lip(\Phi(\x_0) - \Phi^\star)}{c\varepsilon^2} \le \delta
\end{equation*}
which finishes the proof.

\end{proof}

\end{longversion}

\paragraph{Remarks on large $\lambda$}
We point out that when $\lambda$ is large enough so that the $g$ term alters the local landscape of the objective function $\Phi(\x)$, it is inevitable that new local minima will be introduced to the landscape of the objective function, and potentially change the stability of saddle points. We hypothesize that perturbed proximal descent will still converge to an \ESSP regardless of the magnitude of $\lambda$.

An example for the new local minima introduced by large $\lambda$ is Fig.~\ref{fig::octopus_10}. We see new wrinkles are introduced to the four legs of the octopus function as $\lambda$ increases from $1$ to $10$. If an iteration starts in the neighborhood of creases, it can converge to the bottom of the creases. 
Fig.~\ref{fig::octopus_100} is an extreme scenario where the original landscape of the octopus function is completely altered to conform to the behavior of $\ell_1$ penalty term.

\subsection{From \ESSP to local minimizers}

\begin{assumption}[Nondegenerate Saddle]\label{as::nondegenerate saddle}
For all stationary points $\x_c$, 
$\exists\, m>0$ such that $\dis \min_{i=1,2,\cdots,d} |\lambda_i(\nabla^2 f(\x_c))| > m > 0$, where $\lambda_i$ are the eigenvalues (not to be confused with the parameter $\lambda$).
\end{assumption}

With this nondegenerate saddle assumption, the main theorem can be strengthened to the following corollary, whose proof is immediate as one sets the $\varepsilon$ value in the main theorem as $m^2/\rho$ and realizes that there is no eigenvalue of $\nabla^2 f$ existing between $-\sqrt{\rho\varepsilon}$ and the first positive eigenvalue.

\begin{corollary}
There exists an absolute constant $c_{\max}$ such that if $f(\cdot)$ satisfies assumptions \ref{as::smooth_Lip}, \ref{as::small_lambda} and \ref{as::nondegenerate saddle},  then for any $\delta>0, \Delta_\Phi \ge \Phi(\x_0) - \Phi^\star$, constant $c \le c_{\max}$, and $\varepsilon = \frac{m^2}{\rho}$, with probability $1-\delta$, the output of $\text{PPD}(\x_0, \Lip, \rho, \varepsilon, c, \delta, \Delta_f)$ will be a local minimizer of $f + \lambda\|\x\|_1$, and terminate in iterations:
\begin{equation*}\mathcal{O}\left(\frac{\Lip(\Phi(\x_0) - \Phi^\star)}{\varepsilon^2}\ln^{4}\left(\frac{d\Lip\Delta_\Phi}{\varepsilon^2\delta}\right) \right)
\end{equation*}
\end{corollary}

\section{Numerical Experiment}
\label{sec::numerics}
\begin{figure}
  \centering
  \begin{subfigure}[b]{0.32\textwidth}
  \includegraphics[width=\textwidth]{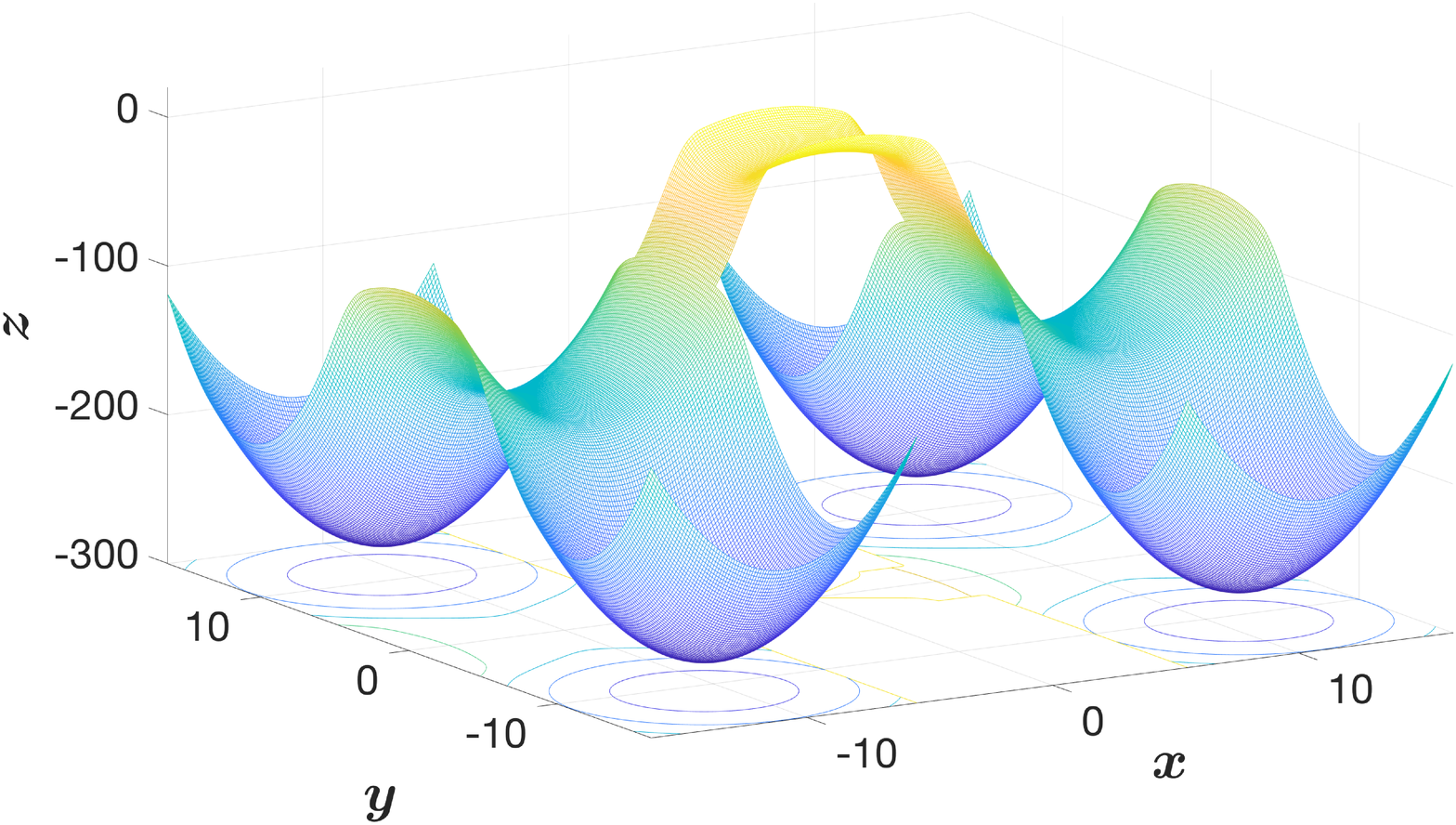}
  \caption{$\lambda=0.01$}
  \label{fig::octopus_.1}
  \end{subfigure}
%
  \begin{subfigure}[b]{0.32\textwidth}
  \includegraphics[width=\textwidth]{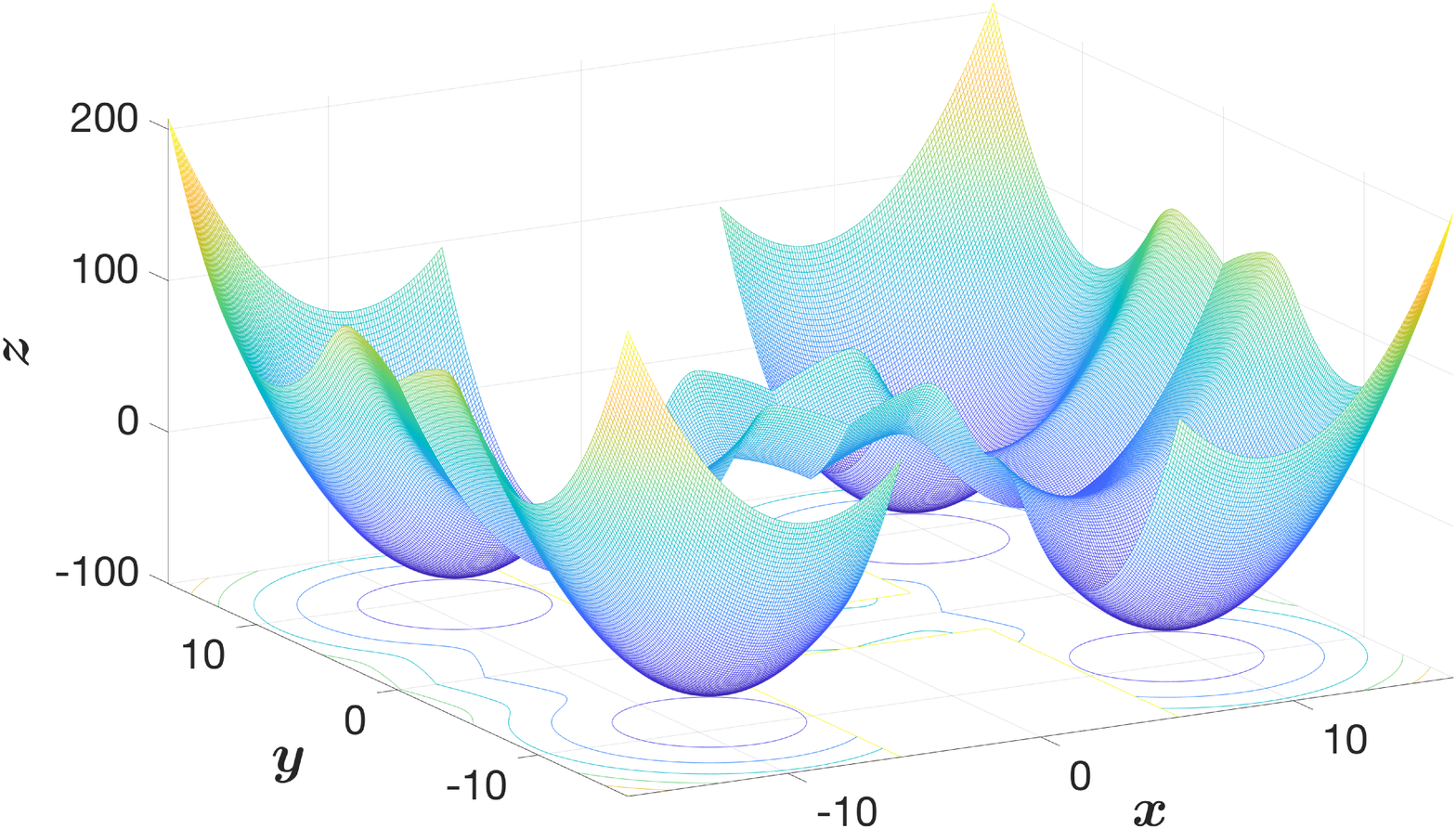}
  \caption{$\lambda=10$}
  \label{fig::octopus_10}
  \end{subfigure}
 \begin{subfigure}[b]{0.32\textwidth}
  \includegraphics[width=\textwidth]{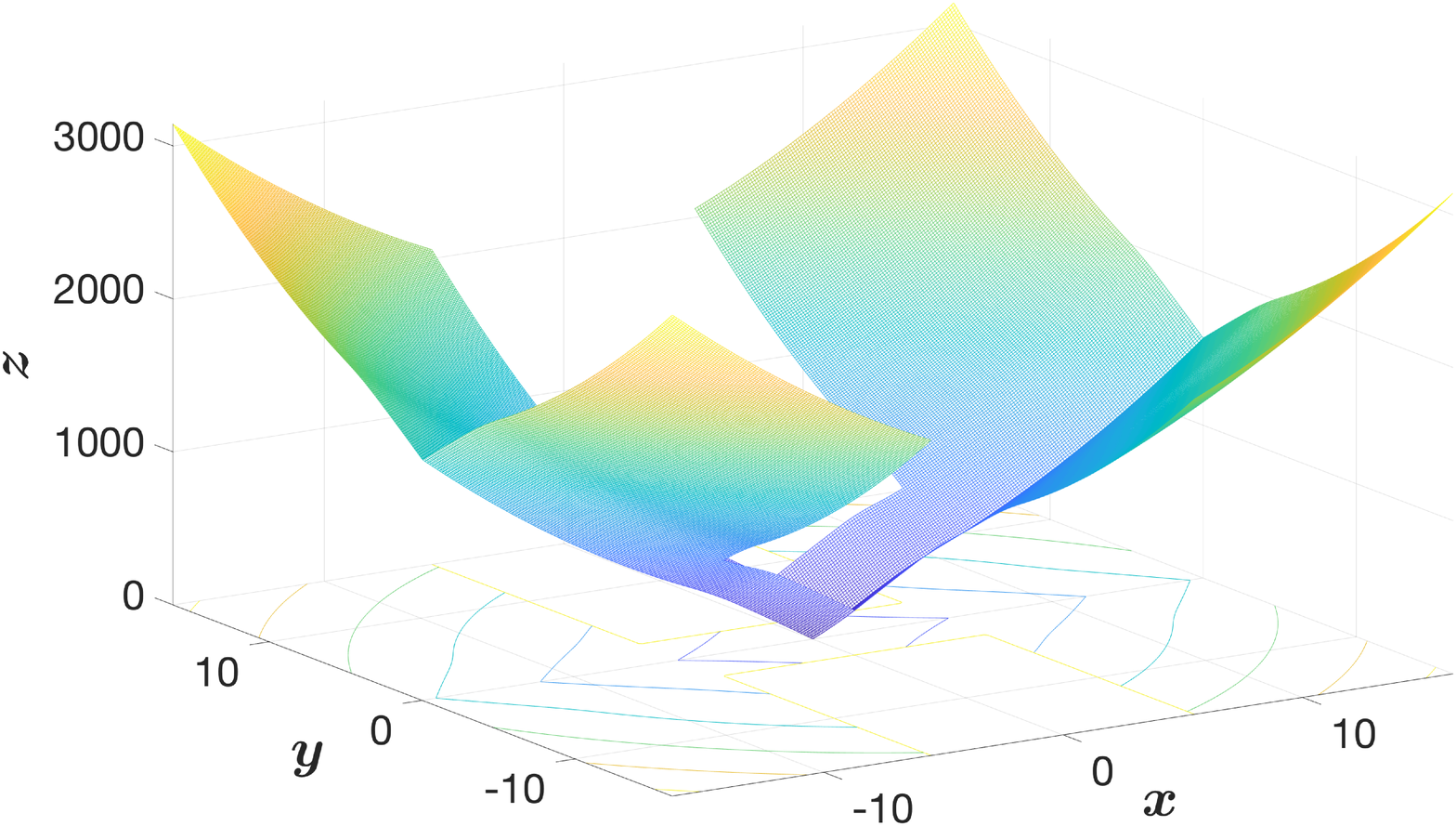}
  \caption{$\lambda=100$}
  \label{fig::octopus_100}
  \end{subfigure}
  
  \caption{The octopus function with different $\lambda$ values}
  \label{fig::octopus}
\end{figure}

We set $f$ to be the ``octopus'' function described in \cite{Du2017Octopus} and use perturbed proximal descent to minimize the objective function $\Phi(\x) = f(\x)+\lambda\|\x\|_1$. Plots of octopus function defined in $\R^2$ for various $\lambda$ are shown in Figure \ref{fig::octopus}.

\newcommand{\OctoL}{M} 
The ``octopus'' family of functions is parameterized by $\tau$, which controls the width of the ``legs," and $\OctoL$ and $\gamma$ which characterize how sharp each side is surrounding a saddle point, related to the Lipschitz constant. The example illustrated in Fig.~\ref{fig::octopus} uses parameters $\OctoL=\mathrm{e},\gamma = 1, \tau = \mathrm{e}$.

We are interested in the octopus family of functions because it can be generalized to any dimension $d$, and it has $d-1$ saddle points (not counting the origin) which are known to slow down standard gradient descent algorithms. The usual minimization iteration sequence, if starting at the maximum value of the octopus function, will successively go through {\it each} saddle point before reaching the global minimum, thus rendering the iteration progress easy to track and visualize.

\begin{longversion}
\paragraph{Specifics of Octopus Function}

We define octopus function in first quadrant of $\R^d$. And then, by even function reflection, the octopus can be continued to all other quadrants.

Define the \textit{auxiliary gluing functions} as
\begin{align*}
\mathcal{G}_1(x_i) &= -\gamma x_i^2 + \frac{-14L + 10\gamma}{3\tau}(x_i - \tau)^3 + \frac{5L-3\gamma}{2\tau}(x_i - \tau)^4\\
\mathcal{G}_2(x_i) &= -\gamma - \frac{10(L+\gamma)}{\tau^3}(x_i-2\tau)^3 - \frac{15(L+\gamma)}{\tau^4}(x_i - 2\tau)^4 - \frac{6(L+\gamma)}{\tau^5}(x_i-2\tau)^5
\end{align*}
Define the \textit{gluing function} and \textit{gluing balance constant} respectively as
\begin{align*}
\mathcal{G}(x_i,x_{i+1}) &= \mathcal{G}_1(x_i)+\mathcal{G}_2(x_i)x_{i+1}^2\\
\nu &= -\mathcal{G}_1(2\tau)+4L\tau^2 = \frac{26L + 2\gamma}{3}\tau^2 + \frac{-5L+3\gamma}{2}\tau^3
\end{align*}
For a given $i=1,\cdots,d-1$, when $6\tau\ge x_1,\cdots,x_{i-1}\ge 2\tau,\tau\ge x_i\ge 0,\tau\ge x_{i+1},\cdots,x_d\ge 0$
\begin{equation}
f(\x) = \sum_{j=1}^{i-1} L(x_j-4\tau)^2 - \gamma x_i^2 + \sum_{j=i+1}^d L x_j^2 - (i-1)\nu\equiv f_{i,1}(\x)
\end{equation}
and if $6\tau\ge x_1,\cdots,x_{i-1}\ge 2\tau,2\tau\ge x_i\ge \tau,\tau\ge x_{i+1},\cdots,x_d\ge 0$, we have
\begin{equation}
f(\x) = \sum_{j=1}^{i-1} L(x_j-4\tau)^2  +\mathcal{G}(x_i,x_{i+1}) + \sum_{j=i+2}^d L x_j^2 - (i-1)\nu\equiv f_{i,2}(\x)
\end{equation}
and for $i=d$, if $6\tau\ge x_1,\cdots,x_{d-1}\ge 2\tau,\tau\ge x_d\ge 0$
\begin{equation}
f(\x) = \sum_{j=1}^{d-1} L(x_j-4\tau)^2 - \gamma x_d^2 - (d-1)\nu\equiv f_{d,1}(\x)
\end{equation}
and if $6\tau\ge x_1,\cdots,x_{d-1}\ge 2\tau,2\tau\ge x_d\ge \tau$
\begin{equation}
f(\x) = \sum_{j=1}^{d-1} L(x_j-4\tau)^2  +\mathcal{G}_1(x_d) - (d-1)\nu\equiv f_{d,2}(\x)
\end{equation}
and if $6\tau\ge x_1,\cdots,x_d\ge 2\tau$,
\begin{equation}
f(\x) = \sum_{j=1}^d L(x_j-4\tau)^2 -d\nu \equiv f_{d+1,1}(\x)
\end{equation}
\end{longversion}

\paragraph{Remark} All saddle points happen at $(\pm4\tau,\pm4\tau,\cdots,\pm4\tau,0,0,\cdots,0)$, and the global minimum is at $(\pm 4\tau,\cdots,\pm4\tau)$. Regions in the form of $[2\tau,6\tau]\times\cdots\times[2\tau,6\tau]\times[\tau,2\tau]\times[0,\tau]\times\cdots\times[0,\tau]$  are transition zones described by the gluing functions which connect separate pieces to make $f$ a continuous function. The octopus function can be constructed first in the first quadrant, and then using even function reflection to define it in all other quadrants. 
A typical descent algorithm applied to the octopus generates iterations that take multiple turns like walking down a spiral staircase, each staircase leading to a new dimension.

\subsection{Results}
\begin{figure}
    \centering
    \includegraphics[width=.9\textwidth]{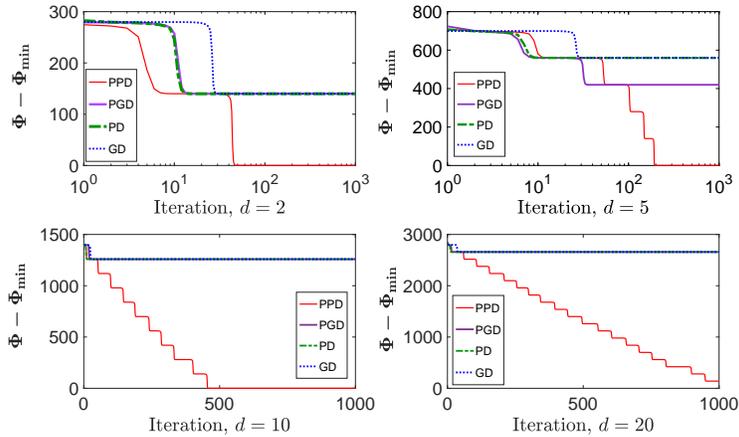}
    \caption{Performance of our proposed PPD algorithm
    on the octopus function with $\lambda=0.01$}
    \label{fig::PPD_demo}
    \vspace{-3mm}
\end{figure}
We  apply the perturbed proximal descent (PPD) on the octopus function plus $0.01\|\x\|_1$ when the dimension varies between $d=2,5,10, 20$. We set the constant $c=3$. 
For comparison, we apply perturbed gradient descent (PGD) as well since $\|\x\|_1$ is differentiable almost everywhere; for both algorithms, the norm of the perturbation $\boldsymbol{\xi}$ is $0.1$. 

We see that PPD successfully finds the local minimum in the first three cases within 1000 iterations, and in the case of $d=20$, PPD almost finds the local minimum within 1000 iterations. In contrast,  unperturbed proximal descent (PD),  gradient descent (GD), and perturbed gradient descent (PGD) sequences are trapped near saddle points.

\section{Conclusion}
This paper provides an algorithm to minimize a non-convex function plus a $\ell_1$ penalty of small magnitude, with a probabilistic guarantee that the returned result is an approximate second-order stationary point,
and hence for a large class of functions, a local minimum instead of a saddle point. The complexity is of $\mathcal{O}(\varepsilon^{-2})$ and the result depends on dimension in $\mathcal{O}(\ln^4 d)$.

The deficiency of the result is that the magnitude of $\ell_1$ penalty needs to be small to let our theoretical result hold. Meanwhile, we also notice that a large $\lambda$ will lead to creation of new local minima to the objective function altering the original landscape. Our future work will address the case of large $\lambda$ in the iteration process.

\bibliography{thesisBecker}{}
\bibliographystyle{plain}

\end{document}